\documentclass[letterpaper, 10 pt, conference]{ieeeconf}
\IEEEoverridecommandlockouts
\usepackage[backend=biber,
            hyperref=true,
            url=false,
            isbn=false,
            doi=false,
            backref=false,
            style=ieee,
            citestyle=numeric-comp,
            sorting=nyt,
            block=none]{biblatex}

\usepackage{flushend}
\usepackage{balance}
\usepackage{xcolor}
\usepackage{multirow}
\usepackage{amsfonts}
\usepackage{siunitx}
\usepackage{textcomp}
\usepackage{graphicx}
\usepackage{xspace}
\usepackage{hyperref}
\usepackage{amsmath}

\newcommand{\methodname}{\textsc{6-PACK}\xspace}

\renewcommand{\baselinestretch}{0.98}

\title{\LARGE \bf
\textsc{6-PACK}: Category-level 6D Pose Tracker with Anchor-Based Keypoints
}

\author{Chen Wang$^{2}$, Roberto Mart\'{i}n-Mart\'{i}n$^{1}$, Danfei Xu$^{1}$, Jun Lv$^{2}$, \\ Cewu Lu$^{2}$, Li Fei-Fei$^{1}$, Silvio Savarese$^{1}$, Yuke Zhu$^{1,3}$ \\
\thanks{$^{1}$Department of Computer Science, Stanford University, USA} \\
\thanks{$^{2}$Department of Computer Science, Shanghai Jiao Tong University, China} \\
\thanks{$^{3}$NVIDIA Research, USA}
}

\begin{document}

\maketitle
\thispagestyle{empty}
\pagestyle{empty}
\vspace{-6mm}

\begin{abstract}
We present \methodname{}, a deep learning approach to category-level 6D object pose tracking on RGB-D data. Our method tracks in real time novel object instances of known object categories such as bowls, laptops, and mugs. \methodname{} learns to compactly represent an object by a handful of 3D keypoints, based on which the interframe motion of an object instance can be estimated through keypoint matching. These keypoints are learned end-to-end without manual supervision in order to be most effective for tracking. Our experiments show that our method substantially outperforms existing methods on the NOCS category-level 6D pose estimation benchmark and supports a physical robot to perform simple vision-based closed-loop manipulation tasks. Our code and video are available at \url{https://sites.google.com/view/6packtracking}.
\end{abstract}

\section{Introduction}

Estimating 6D pose of objects, i.e., translation and orientation in 3D, offers a concise and informative form of state representation for robotic applications, such as manipulation~\cite{GarciaCifuentes.RAL,schmidt2014dart,kappler2018real} and navigation~\cite{xu2017pointfusion,qi2018frustum,3dop}. In robotic manipulation, the ability of tracking object 6D poses in real-time gives rise to fast feedback control~\cite{kappler2018real}. Pioneering work in 6D tracking~\cite{azad20116,schmidt2014dart} has achieved remarkable accuracy and robustness given the 3D model of an object instance, often referred as \emph{instance-level 6D tracking}. However, the assumption of known 3D model can be brittle in realistic settings, where perfect geometry of novel objects is hard to acquire. In this work, we propose to study the problem of \emph{category-level 6D tracking}, where the goal is to develop category-level models capable of tracking novel object instances within a specific category.

The problem of \emph{category-level tracking} has been studied extensively in 2D domains. Classical methods rely on handcrafted features as object representations for visual tracking~\cite{yilmaz2006object,kalal2011tracking,henriques2014high}. Recent work has embarked on an exploration of new computational tools, in particular, deep neural networks, and large amounts of training data to improve tracking performance under visual variations and heavy occlusions~\cite{ondruska2016deep,held2016learning}. However, a model for 6D pose tracking would have to handle the larger search space of all possible poses due to the increased dimensionality, leading to a substantial computational burden over 2D visual trackers.

One remedy is to reduce category-level 6D tracking to a 3D detection and 6D pose estimation problem. 3D detection and pose estimation have been studied in a large body of literature, especially in the context of autonomous driving~\cite{qi2018frustum,xu2017pointfusion}. Most relevant to us is NOCS~\cite{wang2019normalized} which introduced a category-level model to estimate the 6D pose of objects from RGB-D images. NOCS transforms every object pixel to a shared coordinate frame as keypoints for pose estimation. However, estimating poses from a large number of crude keypoints makes their method susceptible to noises from clutter and occlusion. Furthermore, these tracking-by-detection methods cannot leverage temporal information from previous frames. In contrast, we seek to develop a tracking model that learns compact and discriminative object representations for robust registration and leverages temporal consistency for efficient search.

\begin{figure}[t]
    \center
    \includegraphics[width=0.98\linewidth]{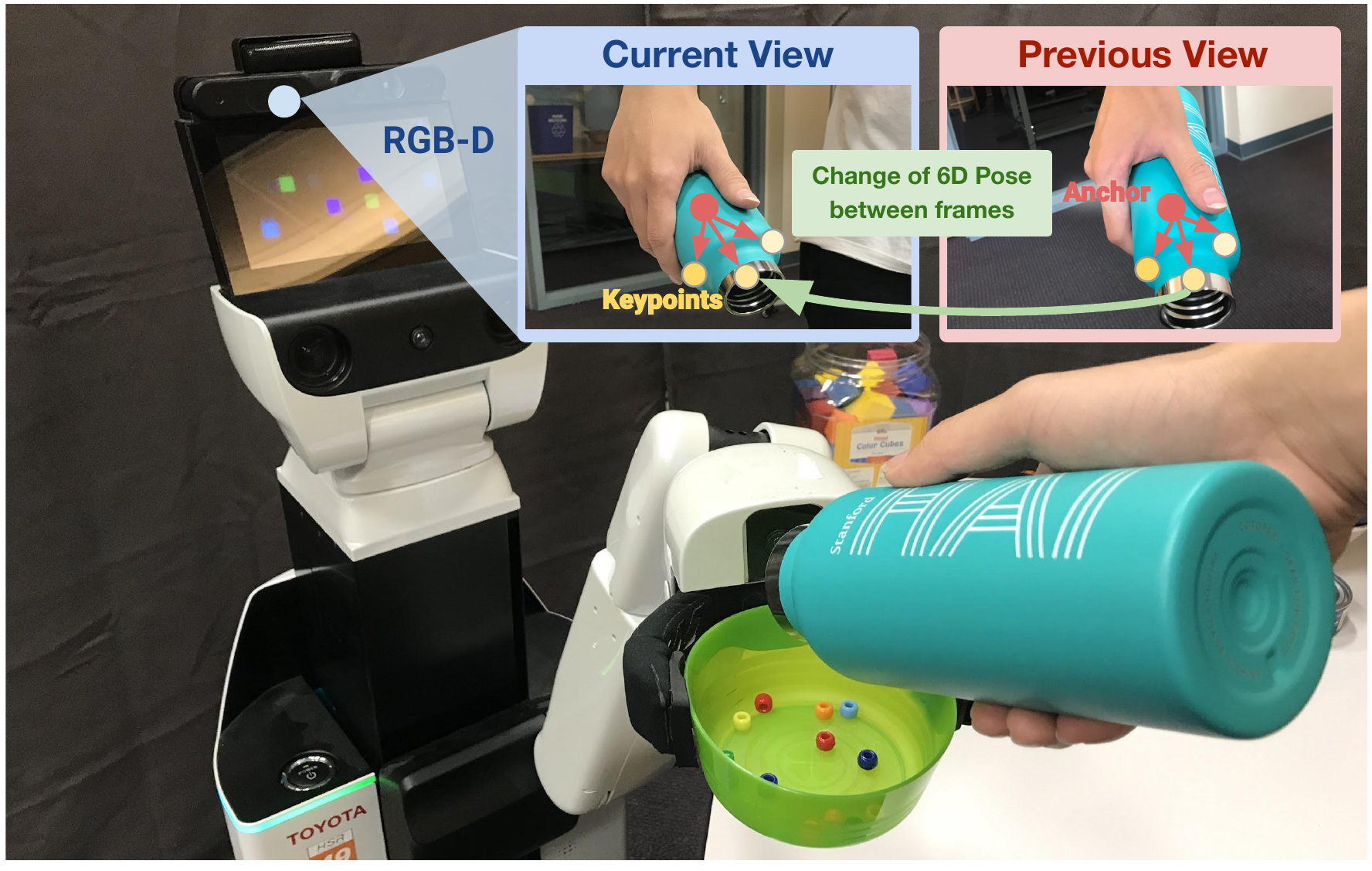}
    \vspace{-2mm}
    \caption{\textbf{Overview of \methodname{}:} Our model learns to robustly detect and track a set of 3D category-based keypoints (yellow dots) based on anchors (red dots) sampled around the previously estimated object pose using RGB-D images. The keypoints from two consecutive frames are then used to compute the 6D object pose change via least-squares optimization. The entire process is fast (\textgreater10fps) to enable real-time robot interaction.}
    \vspace{-5mm}
    \label{fig:pullfig}
\end{figure}

To this end, we propose \methodname, a vision-based \textbf{6}D-\textbf{P}ose \textbf{A}nchor-based \textbf{C}ategory-level \textbf{K}eypoint tracker. \methodname tracks a small set of keypoints in RGB-D videos and estimates object pose by accumulating relative pose changes over time (see Fig.~\ref{fig:pullfig}). This method does not require the known 3D model. Instead, it circumvents the need of defining and estimating the absolute 6D pose via a novel anchor mechanism analogous to the proposal methodology used in 2D object detection~\cite{ren2015faster}. These anchors offer a base for generating 3D keypoints. Unlike previous methods that require manual keypoint annotations~\cite{manuelli2019kpam}, we propose an unsupervised learning approach that discovers the optimal set of 3D keypoints for tracking. These keypoints serve as a compact representation of the object, from which its pose difference between two adjacent frames can be efficiently estimated~\cite{arun1987least}. This keypoint-based representation leads to robust and real-time 6D pose tracking.

We show experimentally improved generalization and robustness in category-level 6D tracking compared to the traditional registration-based tracker~\cite{segal2009generalized}, the state-of-the-art tracking-by-detection method~\cite{wang2019normalized}, and other ablative baselines. \methodname{} substantially outperforms all baselines on the recently introduced NOCS-REAL275 dataset~\cite{wang2019normalized}. The proposed tracker runs at 10Hz on a GTX1070 GPU. We deploy this tracker on a Toyota HSR Robot and demonstrate its utility in real-time manipulation tasks.

\section{Related Work}
\label{s:rw}

Historically, 6D object visual pose estimation has relied on matching the current view of an object to a given template model. The pose can then be recovered from solving an error minimization problem between correspondences, either a PnP, a reprojection or a distance error, depending on the nature of the template~\cite{collet2011moped, choi2010real, lepetit2004point, lepetit2006keypoint}. While conceptually simple and efficient in practice, the performances of these methods degrade significantly under clutter or variable lighting conditions due to errors in feature matching. 

Recent 6D object pose estimation methods instead learns to directly match input images with renderings~\cite{li2018deepim,deng2019poserbpf} or silhouette~\cite{hu2019segmentation,billings2018silhonet} of template object models. PoseRBPF~\cite{deng2019poserbpf} compares the latent code of the input image and that of the model rendering to recover the rotation part of the object pose. However, the requirements of knowing the object models confine these methods to known object instances. 

The problem of \textit{category-level} pose estimation has amassed significant interests in research areas such as autonomous driving~\cite{xu2017pointfusion,qi2018frustum,zhou2017voxelnet} due to the availability of large-scale datasets~\cite{kitti}. Most relevant to us is a class of methods that combines the classical keypoint matching ideas and modern learning techniques by directly predicting either category-level semantic keypoints~\cite{tulsiani2015viewpoints, pavlakos20176} or 3D bounding box corners~\cite{law2019cornernet,tekin18,tremblay2018deep}. However, supervised keypoints learning require large amounts of labelled data, and the manually annotated keypoints or the bounding box corners may not be the optimal landmarks to track. Moreover, defining features \textit{a priori} (so-called \textit{feature engineering}) has largely been outperformed by data-driven methods that learn the best features for the task~\cite{bengio2013representation}.

A notable exception is NOCS~\cite{wang2019normalized}, which learns to project every input object pixel into a category-level canonical 3D space as keypoint. The pose is then recovered by registering all projected keypoints to an ``average'' category object model. However, as we show experimentally, the reliance on dense keypoint correspondences makes NOCS susceptible to noise from occlusion. In contrast, our novel keypoint detector is trained end-to-end only with the final pose tracking objective to generate a small but robust set of 3D keypoints for tracking without direct keypoint supervision.

Our keypoint generator is closely related to the large corpus of keypoint detection and matching methods. Classical methods detect and match keypoints based on hand-designed local features~\cite{lowe2004distinctive, rublee2011orb, bay2006surf, rosten2006machine, calonder2010brief}. Modern learning approaches learns to detect keypoints via optimizing fully-supervised \cite{huang2017coarse,chen2019adversarial,alp2018densepose} or semi-supervised objectives~\cite{honari2018improving}. Recently, KeypointNet~\cite{suwajanakorn2018discovery} shows the benefits of learning to generate keypoints in 3D space \textbf{without supervision} by exploiting geometric consistency across multiple views. The model generalizes to new object instances within known categories in a synthetic domain. However, as we show in our experimental evaluation, KeypointNet fails to scale to a real-world since the model has been trained to define keypoints on single models of objects centered at the origin of coordinates. This approach suffers when exposed to noisy RGB-D data from a real scene where the objects have occlusions and are not centered, increasing the boundaries of the 3D space where the keypoints could be found. Our method introduces a novel anchoring mechanism that allows the model to generate keypoints only in the most relevant subspace. The strategy significantly improves the quality of the generated keypoints and reduces the number of keypoints needed for estimating the pose, enabling our model to track objects in real-time.

\section{Problem Definition}
\label{s:back}

For any instance of an object category, given its initial 6D pose, $p_{0}\in \mathbf{SE}(3)$, and the category it belongs to (e.g., \texttt{mug}, \texttt{bowl}, \texttt{laptop}, \ldots), we define \textbf{category-level 6D pose tracking} as the problem of estimating continuously the change of pose of the object between consecutive timesteps $t-1$ and $t$, $\Delta p_{t}\in \mathbf{SE}(3)$. A change of pose consists of change in rotation $\Delta R_{t} \in \mathbf{SO}(3)$ and a change in translation $\Delta t_{t} \in \mathbb{R}^3$, $\Delta p_{t}=[\Delta R_{t}|\Delta t_{t}]$. The absolute pose can be then retrieved by applying recursively the last estimated change of pose: $p_t = \Delta p_{t}\cdot p_{t-1} = \Delta p_{t}\cdot\Delta p_{t-1} \cdots p_{0}$. 

The initial pose is the translation and rotation with respect to the camera frame of a canonical frame defined similarly for all instances of the same category. 
This setup was defined in NOCS~\cite{wang2019normalized} for the related problem of category-level 6D pose estimation. 
For example, for the category \texttt{camera}, the frame is placed at the centroid of the object with the $x$-axis pointing in the direction of the camera objective and the $y$-axis pointing upwards. Similar to prior 6D tracking work~\cite{wuthrich2013probabilistic, issac2016depth}, we assume the initial pose of the object is given. 
However, different from these approaches, our method is robust to errors in this initial pose, as shown in our experimental evaluation (Sec.~\ref{s:exp}).

Our method is inspired by recent works on learning to predict 3D keypoints for 6D object pose estimation~\cite{tulsiani2015viewpoints, pavlakos20176,suwajanakorn2018discovery}. Following~\cite{suwajanakorn2018discovery}, we define 3D keypoints as points in 3D space $k \in \mathbb{R}^3$ that are geometrically and semantically consistent throughout a temporal sequence. Fig.~\ref{fig:pullfig} illustrates an example of such keypoints. Concretely, given two consecutive input frames $(I_{t-1}$, $I_{t})$, the problem is to predict matching keypoint lists $(\mathbf{k}_{t-1}$, $\mathbf{k}_{t})$ from the two frames. Then based on rigid body assumption, the change of pose $\Delta p_t$ can be recovered by solving a point set alignment problem with least-squares optimization~\cite{arun1987least}.

\begin{figure*}[ht]
    \center
    \includegraphics[width=0.96\linewidth]{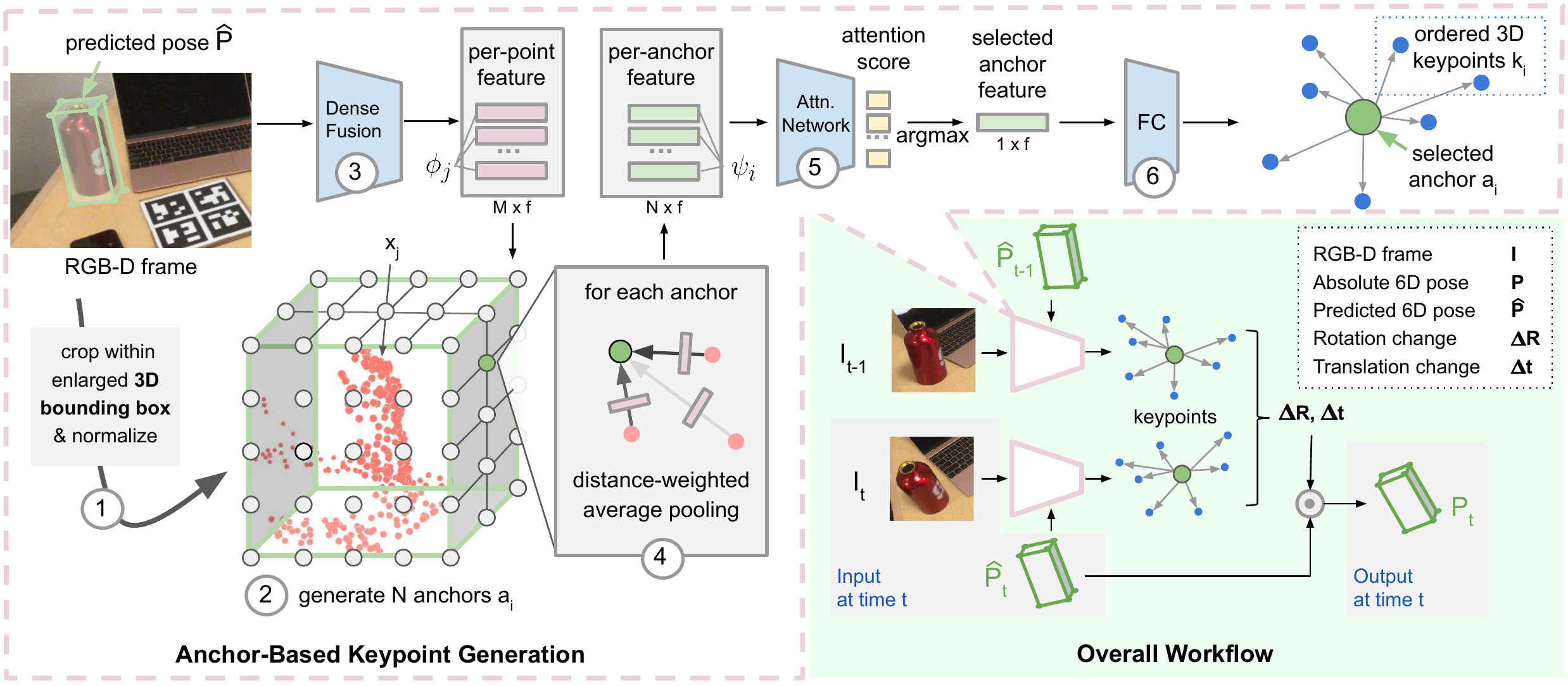}
    \vspace{-3mm}
    \caption{\textbf{Anchor-Based Keypoint Generation:} First, we (1) crop an enlarged volume around the predicted pose of the instance, normalize to a unit, and (2) generate a grid of anchor points on the volume. Then, we (3) use a DenseFusion-based~\cite{wang2019densefusion} network to compute fused geometric and color features for the $M$ points in the volume, and (4) average-pool them into $N$ anchor features based on their distance. The anchor features are (5) used by an attention network to select the closest to the centroid, which is then used to (6) generate an ordered set of keypoints. \textbf{Overall Workflow (bottom right):} The anchor-based keypoint generation is applied on the previous and current frames to obtain two sets of ordered keypoints to compute the inter-frame change in pose. This pose is applied to the previously computed instance pose to obtain the current estimated 6D pose.}
    \label{fig:overall}
    \vspace{-4mm}
\end{figure*}

\section{Model}
\label{s:mod}

\methodname{} performs category-level 6D pose tracking in the following manner (Fig.~\ref{fig:overall}, bottom-right). {First}, \methodname{} uses an attention mechanism over a grid of anchor points (Sec.~\ref{ss:abam}) generated around the predicted pose of the object. Each anchor summarizes the volume around it with a distance-weighted sum of the individual features of the RGB-D points in its surrounding. This information allows to find a coarse centroid of the object in the new RGB-D frame and guide the following search of keypoints around it, which is more efficient than searching keypoints in the entire unconstrained 3D space of previous approaches~\cite{suwajanakorn2018discovery}. 

{Second}, \methodname{} uses the anchor feature to generate keypoints for both symmetric and non-symmetric categories (Sec.~\ref{ss:ukg} and Fig.~\ref{fig:overall} left and top-right). Different from previous methods (e.g., kPAM~\cite{manuelli2019kpam}), these keypoints are learned in an unsupervised manner so they are the most robust and informative for tracking based on training data.

{Finally}, the keypoints of the current and previous frames are passed to a least-squares optimization~\cite{arun1987least} that calculates the inter-frame change in pose. Based on this motion, we extrapolate the pose in the next frame to center the next distribution of anchor points.

The process starts at the location indicated by a given initial pose, $p_0$. We allow for this initial pose to contain an error that we reject with an initial iterative procedure. We generate a set of keypoints and refine the given pose to be at the centroid of the set, $p'_0$. The centroid of the keypoints is close to the instance centroid as it is imposed in the training process (Sec.~\ref{ss:ukg}). We then run the generation and correction again, a total number of $T=10$ times. We select the refined pose that is closest to the centroid of generated keypoints as initial pose for the tracking procedure, since this refined pose is most likely to be the correct one for the instance of the class. The initialization process reduces the effort of providing a very accurate initial pose $p_0$ and increases the robustness of the category-level tracker.

\subsection{Anchor-based Attention Mechanism}
\label{ss:abam}

Directly generating a set of ordered 3D keypoints for pose tracking is challenging due to the large output space. Prior work on automatic keypoint generation~\cite{suwajanakorn2018discovery} did not address this problem: in their setup, keypoints are generated for a single object within a bounded, origin-centered sphere. However, in our setup, the object can be anywhere in 3D space within the field of view of the RGB-D frame. Thus, the keypoints can be anywhere in the unbounded 3D space. 

The stated problem of 3D keypoint generation in unbounded space resembles that of 2D object detection, where the goal is to draw a tight bounding box around the target object on the 2D image. A successful solution to this problem is to begin by drawing a grid of anchor points in the image and find the anchor that is closest to the object center. This procedure locates coarsely the object and simplifies the second step, the generation of a more accurate bounding box proposal around the anchor~\cite{redmon2016you}. Inspired by this idea, we propose an attention mechanism over a grid of 3D anchors around the predicted current object location. Each anchor contains a feature representation of the surrounding volume around it. The model learns to attend, based on this feature, to the anchor closest to the centroid of the object. The 3D keypoints could then be generated as offset points from the selected anchor (Sec.~\ref{ss:ukg}). By splitting the tracking problem into coarse attention-based anchor selection and fine-grained keypoint generation, our tracker has potential to tackle larger region of search space (improving robustness and perceivable inter-frame motion) while maintaining high tracking quality.

As mention before, each anchor contains a feature representing the 3D points in the volume surrounding it. As showed by prior work~\cite{wang2019densefusion}, it is possible to combine color and geometric information into a fused feature to be used for pose estimation. Therefore, we apply the DenseFusion~\cite{wang2019densefusion} feature embedding to all colored 3D points from the RGB-D image within the grid of anchors, and use a distance-weighted averaging to pool them into each anchor to generate the anchor embedding.

Let's assume the grid has $N$ anchor points, $a_i\in\mathbb{R}^3$, and contains $M$ colored points from the RGB-D frame, $x_j\in\mathbb{R}^3$. The vector of distances from an anchor point to all colored points is thus $\mathbf{d}_i = [d_{i0}, d_{i1}, \cdots, d_{iM}]$. Then, the weights for the features of the colored points are defined as $\mathbf{w} = \text{softmax}(\mathbf{d}_i)$. Based on these weights we perform a distance-weighted average pooling to generate the anchor embedding $\psi_i$ of the anchor $a_i$, $\psi_i = \sum_{j}\mathbf{w}_j \phi_j$
, where $\phi_j$ is the point embedding of $x_j$ generated by the DenseFusion encoder. 

Once we have generated a per-anchor geometric and color feature, we train an attention network that learns to detect the one closest to the object centroid. Our attention network takes as input each anchor-level embedding $\psi_i$ and produces as output a confidence score $c_i$ per anchor. The attention network is trained to assign the highest confidence score to the anchor closest to the centroid of the object with supervision. Thus, given the ground-truth position of the object centroid $o_{gt}$, the loss function can be written as:

\begin{equation}
L_{\textit{anc}} = \frac{1}{N}\sum_{i} c_i (||a_i - o_{gt}||_2 - \beta)
\end{equation}
where $\beta = \min(||a_i - o_{gt}||_2), i=1\ldots N$ refers to the minimum possible distance from the anchor to the object centroid. We use a two-layer MLP as the attention network. During the evaluation, \methodname{} selects the anchor with the highest confidence score and generates the keypoints as offsets to this anchor, as explained in the next section. 

\subsection{Unsupervised 3D Keypoints Generation}
\label{ss:ukg}

With the anchor-based attention mechanism, \methodname{} identified the anchor $a_i$ and associated feature $\psi_i$ with the highest confidence score. Now, \methodname{} will use this feature to generate the final set of 3D keypoints, $[k_0,\ldots,k_{K}]$, to track the instance of the object category. We propose a keypoint generation neural network that uses as input the anchor feature and generates a $K\times3$ dimensional output containing an \emph{ordered list} of keypoints. Since the list is ordered, we do not need to find correspondences between keypoints of consecutive frames to estimate the change in pose.

As mentioned before, we train our keypoint generation network in an unsupervised manner that does not require of manual annotation, and that, compared to supervised methods, has led to improved transferability on category-level orientation estimation~\cite{suwajanakorn2018discovery}. We render the unsupervised training of our keypoint generation network as optimizing the multi-view consistency between the keypoints generated in consecutive frames. In other words, suppose $K$ keypoints are generated in each of two consecutive frames; the training objective is to place the keypoints in the current view at the location that corresponds to the keypoints of the previous frame, transformed by the ground truth inter-frame motion. This objective can be formalized in the following \textbf{multi-view consistency loss}:

\begin{equation}
L_\textit{mvc} = \frac{1}{K}\sum_{i}||k_i^{t} - [\Delta R^\textit{gt}_{t}|\Delta t^\textit{gt}_{t}] \cdot k_i^{t-1}||
\end{equation}
where $[\Delta R^\textit{gt}_{t}|\Delta t^\textit{gt}_{t}]=\Delta p^\textit{gt}_{t}$ is the ground truth inter-frame change in pose. 

The multi-view consistency loss only guarantees inter-frame consistency between features locations independently of the perspective or the visible part of the object. However, this does not guarantee these locations are optimal for our final goal, to estimate the change of pose (e.g., all keypoints could end up all at the same location). To tackle this problem we turn the keypoint-based pose estimation step into a differentiable \textbf{pose estimation loss} function and combine it with the multi-view consistency loss in the training process~\cite{suwajanakorn2018discovery}. This loss function consists of a translation loss $L_\textit{tra}$ and rotation loss $L_\textit{rot}$:

\begin{equation}
L_\textit{tra} = ||(\bar{k}^{t} - \bar{k}^{t-1}) - \Delta t^\textit{gt}_{t}||
\end{equation}
\begin{equation}
L_\textit{rot} = 2\arcsin(\frac{1}{2\sqrt{2}}||\Delta\hat{R}_t - \Delta R^\textit{gt}_{t}||)
\end{equation}
where $\bar{k}^{t}$ and $\bar{k}^{t-1}$ are the centroids of the keypoints in previous and current frames, and $\Delta\hat{R}_t$ is the inter-frame change in orientation estimated based on the generated keypoint sets using least-squares optimization~\cite{arun1987least}. Thus, these losses force the keypoints to be generated such that the ground truth change of pose can be computed from them. 

We also integrate a \textbf{separation loss} $L_\textit{sep}$ and \textbf{silhouette consistency loss} $L_\textit{sil}$ defined by Suwajanakorn et al.~\cite{suwajanakorn2018discovery}. The separation loss forces keypoints to maintain some distance to each other to avoid degenerate configurations and improve the pose estimation. The silhouette consistency loss forces keypoints to be closer to the object's surface to improve interpretability.

In addition to the main objectives introduced above, we impose a \textbf{centroid loss} that forces the centroid of the generated set of keypoints to be at the centroid of the object, $L_\textit{cen}$, useful to correct for noise in the given initial pose as explained at the beginning of this section. The final overall training loss is a weighted sum of the 6 terms introduced above, where the weightings are determined by their relative magnitude and importance.

\begin{figure}[t]
\center
\includegraphics[width=0.97\linewidth]{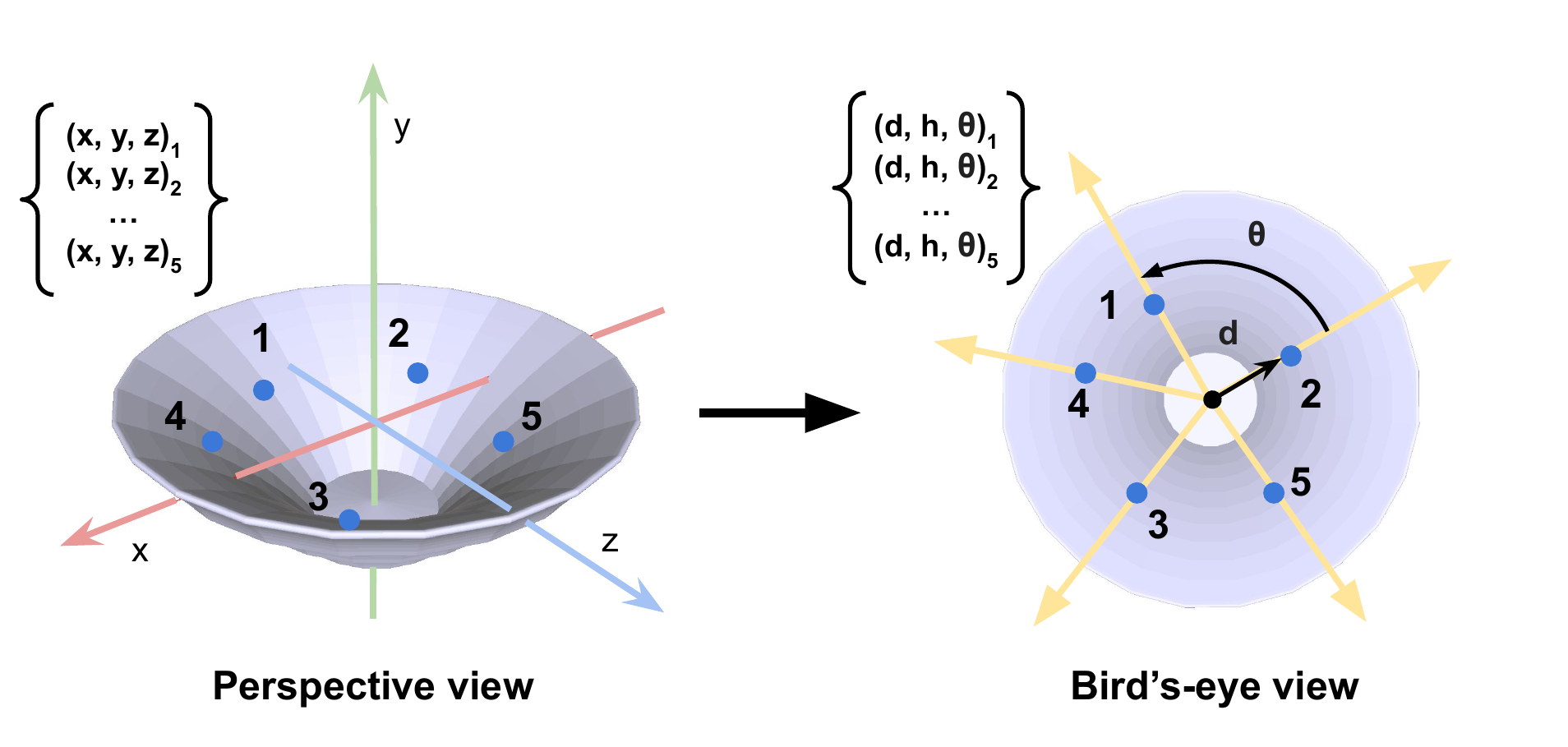}
\label{fig:symmetry}
\vspace{-3mm}
\caption{\textbf{Object Classes with Symmetry Axes:} An instance of the class \texttt{bowl} with an axis of symmetry along \textbf{y}. \textit{Left:} perspective view of the object and Cartesian coordinates of five keypoints. \textit{Right:} Bird's eye view of the object and same points transformed into the symmetry-invariant coordinates (\textbf{d} and \textbf{$\theta$} indicated for keypoint 2).}
\vspace{-4mm}
\end{figure}

\textbf{3D Keypoint Generation for Classes with Symmetry Axes:} The presented multi-view consistency and pose loss functions do not handle well symmetries on instances of object categories since identifying the rotation along the symmetry axis is an unsolvable problem. We propose a coordinate system transformation $\rho()$ that transforms the coordinates of points into a space that is rotation-invariant around the axis of symmetry. Fig.~\ref{fig:symmetry} illustrates the transformation for a case with five points on an instance of \texttt{bowl}. Suppose some common axis of symmetry for all instances in this category, ${s}_\textit{gt}$, passing through the Cartesian origin of coordinates (\textbf{y}-axis in Fig.~\ref{fig:symmetry}); we transform the location of a keypoint point $k_i$ from $(x,y,z)$ in the Cartesian coordinates to the triplet $(d,h,\theta)$ defined as:
\begin{itemize}
    \item \textbf{Distance to the symmetry axis $d$}: The distance from $k_i$ point to the symmetry axis, ${s}_\textit{gt}$.
    \item \textbf{Height along the symmetry axis $h$}: Distance between the orthogonal projection of $k_i$ onto the symmetry axis and the Cartesian origin of coordinates.
    \item \textbf{Relative angle $\theta$}: The separating angle between the radial vector connecting the point $k_i$ to the symmetry axis and the radial vector of the next keypoint encountered when advancing clockwise around the ${s}_\textit{gt}$.
\end{itemize}

The new coordinates are closely related to cylindrical coordinates, where we replace the absolute rotation angle by a relative inter-keypoint angle. Based on the symmetry-invariant transformation, we redefine the multi-view consistency loss for symmetric categories as:

\begin{equation}
L^\textit{sym}_\textit{mvc} = \frac{1}{K}\sum_{i}||\rho(k_i^{t}) - \rho([\Delta R^\textit{gt}_{t}|\Delta t^\textit{gt}_{t}] k_i^{t-1})||
\end{equation}

The pose estimation loss also needs to be adapted for categories with symmetry axes. While the translation loss remains unaltered, we redefine the rotation loss as the angular difference between the predicted change $\Delta \hat{s}$ and the ground-truth change $\Delta {s}_\textit{gt}^{t}$ in orientation of the symmetry axis. The rotation loss for these categories is then simply:
\begin{equation}
L_{R}^{sym} = \arccos\left(\frac{\Delta {s}_\textit{gt}^{t} \cdot \Delta \hat{s}^{t}}{||\Delta {s}_\textit{gt}^{t}||\cdot||\Delta \hat{s}^{t}||}\right)
\end{equation}

\begin{table}[htb]
\centering
\caption{Quantitative evaluation of 6D Pose on NOCS-REAL275}
\vspace{-2mm}
\begin{tabular}{|l|l|c|c|c|c|c|}
\hline
\multicolumn{2}{|l|}{} & \shortstack{NOCS\\\cite{wang2019normalized}} & \shortstack{ICP\\\cite{Zhou2018}} & \shortstack{Keypoint\\Net~\cite{suwajanakorn2018discovery}} & \shortstack{\\Ours w/o \\temporal} & \shortstack{Ours} \\ \hline\hline
\multirow{4}{*}{\texttt{bottle}} & 5\textdegree 5cm & 5.5 & 10.1 & 5.9 & 23.7 & \textbf{24.5} \\
 & IoU25 & 48.7 & 29.9 & 23.1 & \textbf{92.0} & 91.1 \\
 & R$_{err}$ & 25.6 & 48.0 & 28.5 & 15.7 & \textbf{15.6} \\
 & T$_{err}$ & 14.4 & 15.7 & 9.5 & 4.2 & \textbf{4.0} \\ \hline
\multirow{4}{*}{\texttt{bowl}} & 5\textdegree 5cm & \textbf{62.2} & 40.3 & 16.8 & 53.0 & 55.0 \\
 & IoU25 & 99.6 & 79.7 & 74.7 & \textbf{100.0} & \textbf{100.0} \\
 & R$_{err}$ & \textbf{4.7} & 19.0 & 9.8 & 5.3 & 5.2 \\
 & T$_{err}$ & \textbf{1.2} & 4.7 & 8.2 & 1.6 & 1.7 \\ \hline
\multirow{4}{*}{\texttt{camera}} & 5\textdegree 5cm & 0.6 & \textbf{12.6} & 1.8 & 8.4 & 10.1 \\
 & IoU25 & 90.6 & 53.1 & 30.9 & \textbf{91.0} & 87.6 \\
 & R$_{err}$ & \textbf{33.8} & 80.5 & 45.2 & 43.9 & 35.7 \\
 & T$_{err}$ & \textbf{3.1} & 12.2 & 8.5 & 5.5 & 5.6 \\ \hline
\multirow{4}{*}{\texttt{can}} & 5\textdegree 5cm & 7.1 & 17.2 & 4.3 & \textbf{25.0} & 22.6 \\
 & IoU25 & 77.0 & 40.5 & 42.6 & 89.9 & \textbf{92.6} \\
 & R$_{err}$ & 16.9 & 47.1 & 28.8 & \textbf{12.5} & 13.9 \\
 & T$_{err}$ & \textbf{4.0} & 9.4 & 13.1 & 5.0 & 4.8 \\ \hline
\multirow{4}{*}{\texttt{laptop}} & 5\textdegree 5cm & 25.5 & 14.8 & 49.2 & 62.4 & \textbf{63.5} \\
 & IoU25 & 94.7 & 50.9 & 94.6 & 97.8 & \textbf{98.1} \\
 & R$_{err}$ & 8.6 & 37.7 & 6.5 & 4.9 & \textbf{4.7} \\
 & T$_{err}$ & \textbf{2.4} & 9.2 & 4.4 & 2.5 & 2.5 \\ \hline
\multirow{4}{*}{\texttt{mug}} & 5\textdegree 5cm & 0.9 & 6.2 & 3.1 & 22.4 & \textbf{24.1} \\ 
 & IoU25 & 82.8 & 27.7 & 52.0 & \textbf{100.0} & 95.2 \\
 & R$_{err}$ & 31.5 & 56.3 & 61.2 & \textbf{20.3} & 21.3 \\
 & T$_{err}$ & 4.0 & 9.2 & 6.7 & \textbf{1.8} & 2.3 \\ \hline\hline
\multirow{4}{*}{Overall} & 5\textdegree 5cm & 17.0 & 16.9 & 13.5 & 32.5 & \textbf{33.3} \\
 & IoU25 & 82.2 & 47.0 & 53.0 & \textbf{95.1} & 94.2 \\
 & R$_{err}$ & 20.2 & 48.1 & 30.0 & 17.1 & \textbf{16.0} \\ 
 & T$_{err}$ & 4.9 & 10.5 & 8.4 & \textbf{3.4} & 3.5 \\ \hline
\end{tabular}
\label{tab:nocs}
\vspace{-4mm}
\end{table}

\section{Experiments}
\label{s:exp}

In this section, we would like to answer the following questions: 1) Does our method indeed generate robust 3D keypoints that are suitable for 6D pose tracking? 2) Does our anchor-based attention mechanism improve the overall tracking performance? 3) How robust is our method against variable levels of noise in pose initialization, and 4) Is our method efficient enough for real-time applications such as closed-loop object manipulation?

To answer questions 1), 2) and 3), we evaluate our method and compare it to multiple baselines on the NOCS-REAL275~\cite{wang2019normalized} dataset, the only real-world benchmark dataset for category-level object 6D pose tracking. To answer question 4), we deploy our model on a real robot platform and test the model with another ten unseen objects and show that the model can be successfully used in a collaborative pouring and a ``toasting'' tasks. 

\textbf{Dataset:}
We use the NOCS-REAL275 dataset, which contains six categories including \texttt{bottle}, \texttt{bowl}, \texttt{camera}, \texttt{can}, \texttt{laptop}, and \texttt{mug}. Three of them are categories with axes of symmetry. The training set consists of 275K discrete frames of synthetic data generated with $1085$ models of instances of the classes from ShapeNetCore~\cite{chang2015shapenet} with random poses, and seven real videos with ground truth poses depicting in total three instances of objects of each category. The testing set has six real videos depicting in total three different (unseen) instances for each object category with 3,200 frames in total.

\textbf{Evaluation Metrics:}
We follow NOCS~\cite{wang2019normalized}  and report the following metrics: 1) \textbf{\ang{5}\SI{5}{\centi\meter}} \cite{lin2014microsoft, li2018deepim}, the percentage of tracking results with orientation error $<$ \ang{5} and translation error $<$ \SI{5}{\centi\meter}, and 2) \textbf{IoU25}~\cite{geiger2013vision}, percentage of volume overlap between the prediction and ground-truth 3D bounding box that is larger than 25$\%$, 
In addition, we include other traditionally used metrics from the tracking community: 3) \textbf{R$_{err}$}, mean of the orientation error in degrees, and 4) \textbf{T$_{err}$}, mean of the translation error in centimeters.

\textbf{Baselines:}
We compare three model variants with baselines to showcase the effectiveness of our design choices:
\begin{itemize}
    \item \textbf{NOCS}~\cite{wang2019normalized}: The state-of-the-art category-level 6D pose estimation method that uses per-pixel prediction.
    \item \textbf{ICP}~\cite{Zhou2018}: The standard point-to-plane ICP algorithm implemented in Open3D.
    \item \textbf{KeypointNet}~\cite{suwajanakorn2018discovery}: The implementation of our model without the anchor-based attention mechanism, which direct generates 3D keypoints in the 3D space.
    \item \textbf{Ours without temporal prediction}: An ablation of \methodname{} where the predicted pose in the next frame is the previous estimated pose.
    \item \textbf{Ours}: \methodname{} where the predicted pose in the next frame extrapolates from the last estimated inter-frame change of pose (constant velocity model).
\end{itemize}

\begin{figure}[tb]
\center
\includegraphics[width=0.85\linewidth]{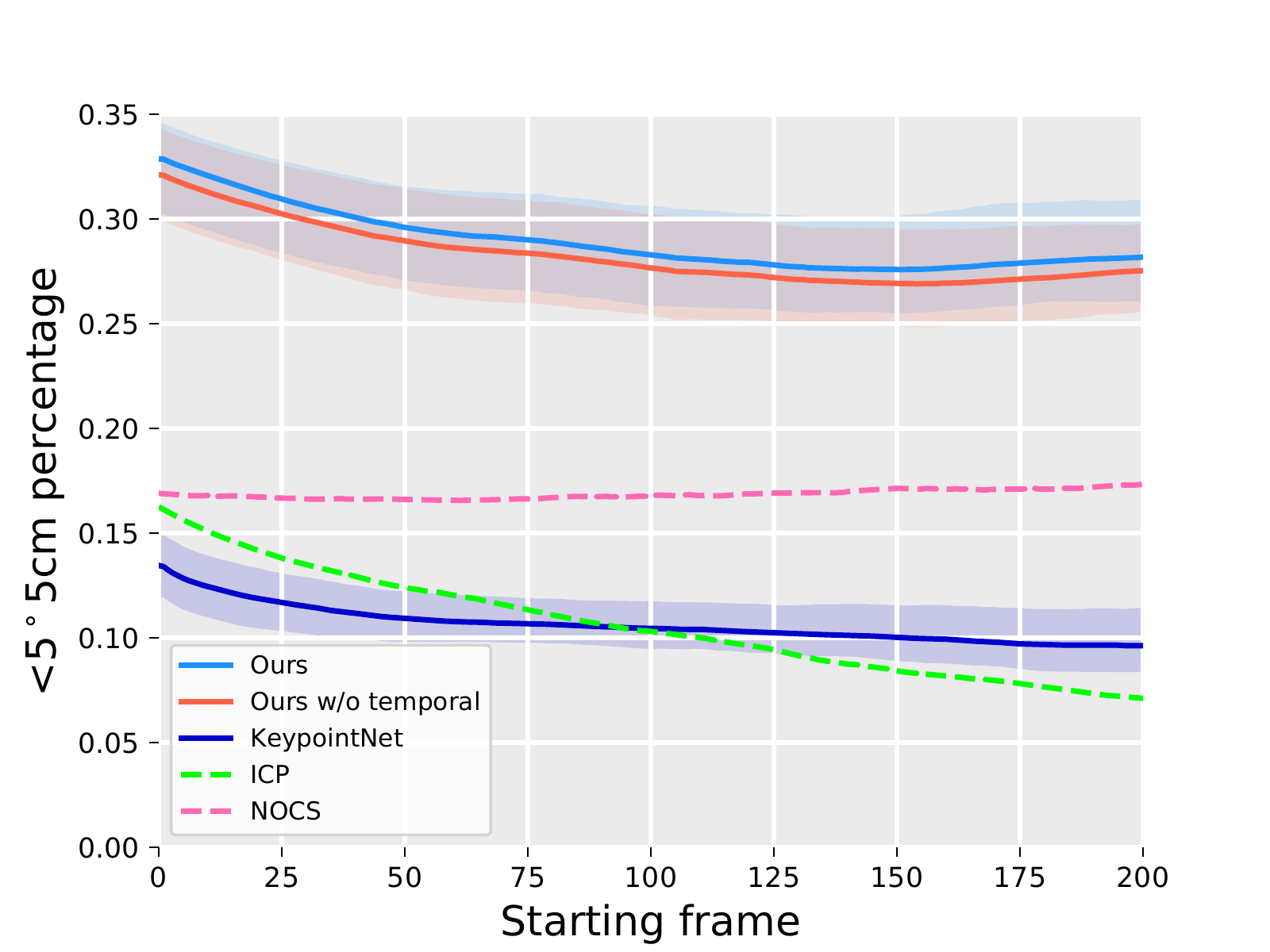}
\vspace{-2mm}
\caption{\textbf{Stability Evaluation over Time}: Each point on the curve represents the mean success rate (\textless \ang{5}\SI{5}{\centi\meter} percentage) on the interval of frames starting from the frame indicated by the \textbf{x}-axis until the end of the sequence, $[x,\text{END}]$. 6-PACK and its variants and KeypointNet are initialized with pose with five different uniformly-sampled translation noise between $\pm$\SI{2}{\centi\meter}. Frames closer to the first frame are easier to estimate since the initial (noisy) pose is given. \methodname{} shows a relatively constant performance even if we do subtract the first easier frames from the computation (drop of only 5\% without the first 75 frames).}
\label{fig:nocs_plot}
\vspace{-3mm}
\end{figure}

\vspace{1mm}
\textbf{Evaluation Results on the NOCS-REAL275 dataset:}
Table~\ref{tab:nocs} contains the results on the testing set of all six categories of the NOCS dataset. We compare two variants of our model (with and without temporal prediction) and several current state-of-the-art category-level 6D pose estimation methods. To evaluate the robustness against noisy initial poses, we inject up to 4cm of uniformly sampled random translation noise. We also measure robustness against missing frames by dropping 450 frames out of 3200 frames are uniformly from the testing videos. 

\begin{figure}[tb]
\center
\includegraphics[width=0.95\linewidth]{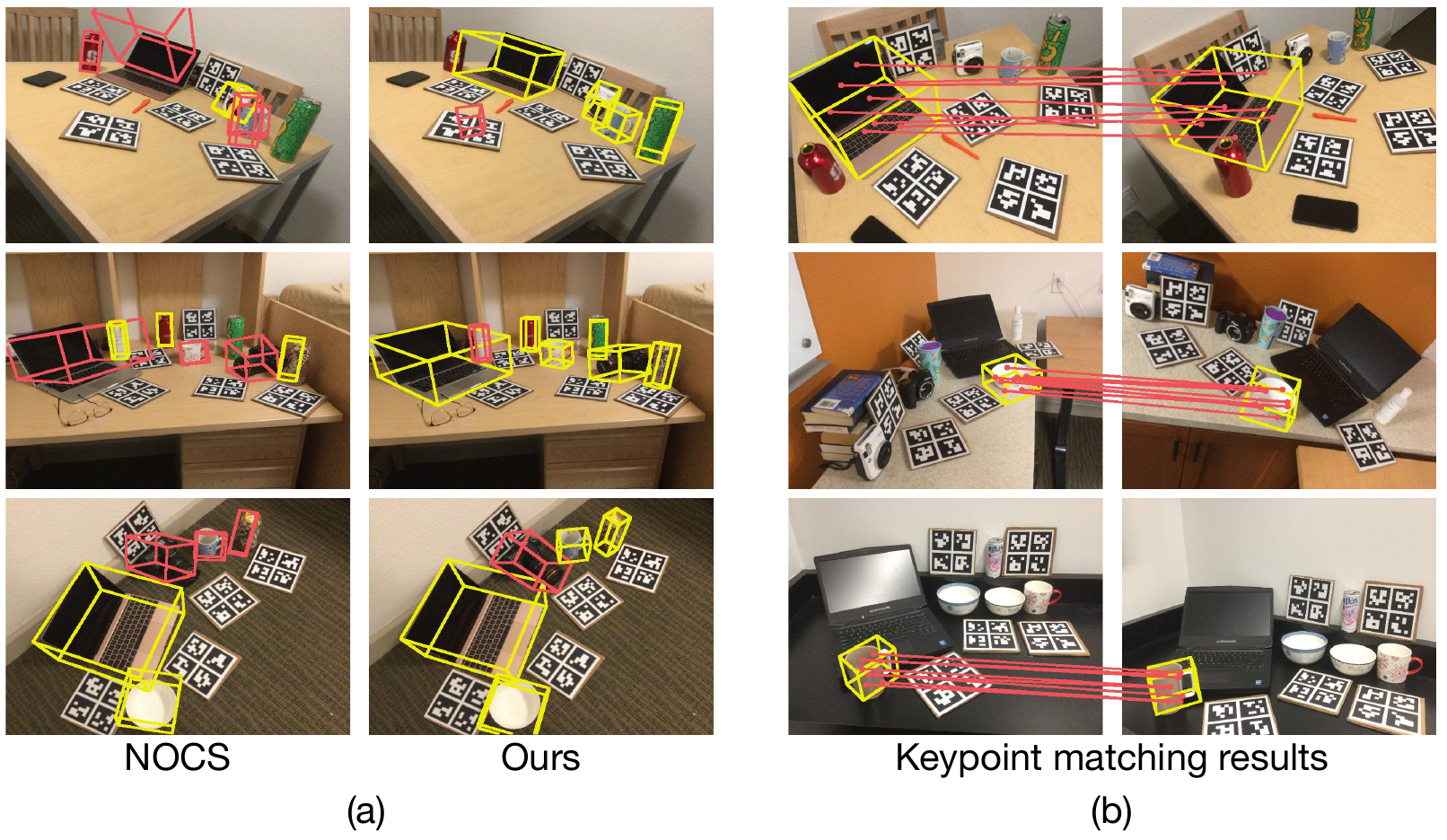}
\vspace{-3mm}
\caption{\textbf{(a) Qualitative Comparison between \methodname{} and NOCS:}  Depicted frames are 100 tracking frames after the initial frame. \textbf{(b) 
Visualization of the Generated Keypoints:} Generated and matched keypoints in two largely separated frames.}
\label{fig:qualitative}
\vspace{-4mm}
\end{figure}

\methodname{} outperforms the second best method NOCS~\cite{wang2019normalized} by more than 15\% in \textbf{\ang{5}\SI{5}{\centi\meter}} metric and 12\% in \textbf{IoU25} metric. Fig.~\ref{fig:qualitative} shows some examples of ours and NOCS pose estimations (left), as well as visualizations of the generated and matched keypoints (right) by \methodname{}. These results indicate that, compared to NOCS, which uses all input pixels as keypoints, our method detects compact and robust 3D keypoints that are best suited for category-based 6D tracking. 

Additionally, \methodname{} outperforms KeypointNet~\cite{suwajanakorn2018discovery} by a large margin in all metrics. The IoU25 metric reveals that KeypointNet frequently loses track of the object (50.3\% across the evaluation set). As discussed in Sec.~\ref{s:rw}, KeypointNet is not designed to handle the large unbounded keypoint generation space of our real-world setting. On the other hand, our method avoids losing track of category instances in evaluation (IoU25 \textgreater 94\%). Our anchor-based mechanism increases the space of search avoiding drifting.

To further examine the robustness and stability of different methods, we compute the mean performance without the first $x$ frames. This way we can measure how much of the performance is due to the ground truth initial pose (frames closer to the initial frame are easy to track). Fig.~\ref{fig:nocs_plot} depicts the mean accuracy (\textless\ang{5}\SI{5}{\centi\meter} percentage). We observe that the performance of all methods decreases, except for NOCS because it is a pose estimation method, not a pose tracker. However, the performance of \methodname{} is more than 10\% higher than the NOCS throughout the whole process and stops decreasing approx. 100 after the initial frame.

We also evaluate the sensitivity of our proposed method in using different number of keypoints on the \texttt{laptop} category. Our model with \textbf{8-keypoints} output achieves 62.4\% in the 5\textdegree 5cm metric, surpassing the \textbf{4-keypoints} 55.2\% and the \textbf{16-keypoints} 48.6\% variants. \textbf{8-keypoints} offer the best trade-off between information compression and redundancy.

\textbf{Integration on a Real Robot:}
Finally, we also deploy our model trained on the NOCS dataset directly to control a real-world robot platform from a computer with an NVIDIA GTX1070 GPU and an Intel Core i7-6700K CPU. The robot used in the experiment is a Toyota HSR (Human Support Robot) equipped with an Asus Xtion RGB-D sensor, a holonomic mobile base, and a two-finger gripper. In this conditions, \methodname{} tracks poses at \SI{10}{\hertz} with less than 30\% of the GPU storage (around 2 GB).

In our tests, we move manually for circa \SI{30}{\second} different instances of the known object categories in front of the robot while the robot tracks them and/or follows them with its gaze. For two of the object categories (\texttt{bowl} and \texttt{bottle}) the robot also performs a manipulation task based on the tracking information (pouring or tossing) at the end of the tracking sequence. To quickly provide an initial coarse pose estimation to initiate tracking, we place and detect a checkerboard that delimits the front face of a 3D bounding box around the target object. We test the tracking performance on \texttt{bowl} (4 instances), \texttt{bottle} (3), \texttt{laptop} (2) and \texttt{can} (2), see video on the project website. Our method successfully tracks without loss the objects in more than 60\% of the trials, achieving high accuracy (object visually within the estimated bounding box). 

\section{Conclusion}
We presented \methodname{}, a category-level 6D object pose tracker. Our tracker is based on a novel anchor-based keypoint generation neural network that detects reliably the same keypoints on different instances of the same category and uses them to estimate the inter-frame change in pose. Our method is trained in an unsupervised manner to allow the network to select the best keypoints for tracking. We compared \methodname{} to 3D geometry methods and deep learning models, showing that our method achieves state-of-the-art performance on a challenging category-based 6D object pose tracking benchmark. Furthermore, we deployed \methodname{} on an HSR robot platform and showed that our method enables real-time tracking and robot interaction.

\section*{Acknowledgement}
This work has been partially supported by JD.com American Technologies Corporation (``JD") under the SAIL-JD AI Research Initiative and by an ONR MURI award (1186514-1-TBCJE). This article solely reflects the opinions and conclusions of its authors and not JD or any entity associated with JD.com. We also want to thank Toyota Research Institute for the Human Support Robot which we used to perform our real robot experiments.

\begin{flushright}
\printbibliography
\end{flushright}
\end{document}